\pdfoutput=1
\documentclass[11pt]{article}
\usepackage{listings}
\usepackage{listingsutf8}
\usepackage{subcaption}
\usepackage{booktabs}
\usepackage{siunitx}
\usepackage[preprint]{acl}

\usepackage{times}
\usepackage{latexsym}
\usepackage{xcolor}
\usepackage{booktabs}
\usepackage[T1]{fontenc}

\usepackage[utf8]{inputenc}

\usepackage{microtype}

\usepackage{inconsolata}

\usepackage{graphicx}
\usepackage{listings}
\usepackage{xcolor}
\lstdefinelanguage{json}{
  basicstyle=\ttfamily\small,
  numbers=left,
  numberstyle=\tiny\color{gray},
  stepnumber=1,
  numbersep=5pt,
  showstringspaces=false,
  breaklines=true,
  frame=single,
  backgroundcolor=\color{gray!5},
  literate=
   *{0}{{{\color{blue}0}}}{1}
    {1}{{{\color{blue}1}}}{1}
    {2}{{{\color{blue}2}}}{1}
    {3}{{{\color{blue}3}}}{1}
    {4}{{{\color{blue}4}}}{1}
    {5}{{{\color{blue}5}}}{1}
    {6}{{{\color{blue}6}}}{1}
    {7}{{{\color{blue}7}}}{1}
    {8}{{{\color{blue}8}}}{1}
    {9}{{{\color{blue}9}}}{1}
    {:}{{{\color{black}:}}}{1}
    {,}{{{\color{black},}}}{1}
    {\{}{{{\color{red}{\{}}}}{1}
    {\}}{{{\color{red}{\}}}}}{1}
    {[}{{{\color{red}[}}}{1}
    {]}{{{\color{red}]}}}{1},
}
\usepackage{url}
\title{Token Level Routing Inference System for Edge Devices\thanks{Demo package available at \url{https://github.com/Jianshu1only/Token-Routing}}}

\author{
  \textbf{Jianshu She\textsuperscript{1}\thanks{Email: \texttt{jianshu.she@mbzuai.ac.ae}}},
  \textbf{Wenhao Zheng\textsuperscript{2}},
  \textbf{Zhengzhong Liu\textsuperscript{1}},
  \textbf{Hongyi Wang\textsuperscript{3}},
  \textbf{Eric Xing\textsuperscript{1}},
\\
  \textbf{Huaxiu Yao\textsuperscript{2}},
  \textbf{Qirong Ho\textsuperscript{1}} \\
\\
  \textsuperscript{1}Mohamed bin Zayed University of Artificial Intelligence (MBZUAI) \\
  \textsuperscript{2}University of North Carolina at Chapel Hill \\
  \textsuperscript{3}Computer Science Department at Rutgers University
}

\begin{document}
\maketitle
\begin{abstract}

The computational complexity of large language model (LLM) inference significantly constrains their deployment efficiency on edge devices. In contrast, small language models offer faster decoding and lower resource consumption but often suffer from degraded response quality and heightened susceptibility to hallucinations. To address this trade-off, collaborative decoding, in which a large model assists in generating critical tokens, has emerged as a promising solution. This paradigm leverages the strengths of both model types by enabling high-quality inference through selective intervention of the large model, while maintaining the speed and efficiency of the smaller model. In this work, we present a novel collaborative decoding inference system that allows small models to perform on-device inference while selectively consulting a cloud-based large model for critical token generation. Remarkably, the system achieves a 60\% performance gain on CommonsenseQA using only a 0.5B model on an M1 MacBook, with under 7\% of tokens generation uploaded to the large model in the cloud.

\end{abstract}

\section{Introduction}
\begin{figure*}[h!]
  \centering
  \includegraphics[width=0.9\textwidth]{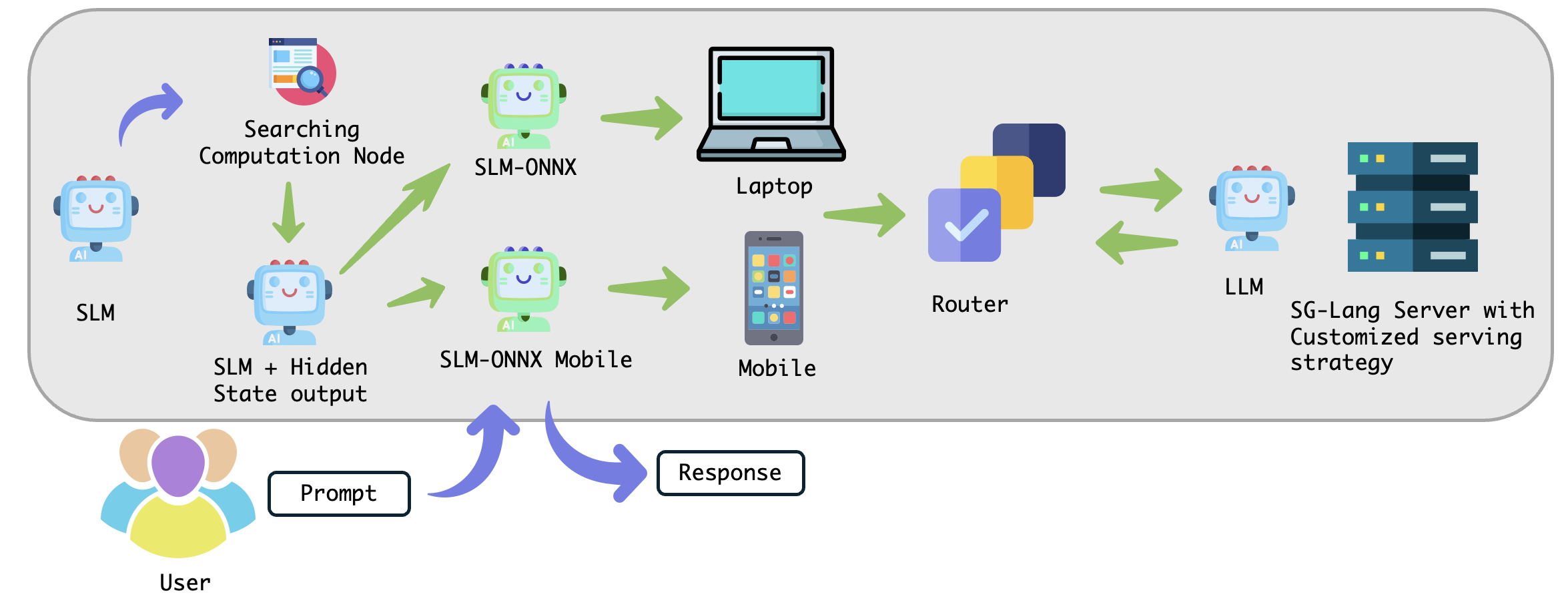}
  \caption{System overview: First transfer Huggingface model to ONNX model, then add hidden states of last layer as a output node in ONNX computation graph, deploy ONNX model on Laptop and ONNX-mobile on Mobile phone. Then connect edge divice with router to the SG-Lang backend from server side. The router automatically route token with low confidence to server, and send response back to edge device}
  \label{fig:system_overview}
\end{figure*}
Large language models (LLMs) have transformed natural language processing, achieving state-of-the-art performance in tasks such as document summarization, question answering, and text generation. Models like Meta's Llama series \cite{touvron2023llamaopenefficientfoundation}, Google's Gemma \cite{gemmateam2024gemmaopenmodelsbased}, and DeepSeek series \cite{deepseekai2025deepseekr1incentivizingreasoningcapability} have demonstrated remarkable capabilities, driving advancements in various applications. However, their deployment in edge devices, such as smartphones, embedded systems, and Internet of Things (IoT) devices, faces significant hurdles due to their high computational complexity 
 \citep{zhang2024fastslowgeneratingempirical,lin2024tinyllm}. The role of small language models (SLMs), and the emerging paradigm of collaborative decoding, culminating in a novel framework that balances efficiency and performance.

The computational demands of LLMs, such as the Llama-2 7B parameter model requiring over 8GB of memory in FP16 precision \cite{zhang2024fastslowgeneratingempirical} , exceed the capabilities of many edge devices, like the NVIDIA Jetson Orin Nano with 8GB DRAM \cite{shen2024learningdecodecollaborativelymultiple,li2025largelanguagemodelinference}. This limitation is compounded by hardware heterogeneity, including ARM processors in smartphones and low-power IoT chips, which further complicates deployment \citep{dao2022flashattentionfastmemoryefficientexact}. Recent works, such as~\citet{zheng2025reviewedgelargelanguage}, highlight the need for solutions that can operate within the constraints of memory, processing power, and energy consumption \citep{Miao_2024}.

One promising approach to leveraging small language models (SLMs) lies in their potential for edge deployment, thanks to their reduced size and faster inference times\citep{xue2024powerinfer2fastlargelanguage,jiang2023llmlinguacompressingpromptsaccelerated,zhou2024surveyefficientinferencelarge}. These models consume fewer resources, making them suitable for devices with limited capabilities. However, studies, such as Wang et al.'s work on large and small model trade-offs \cite{zheng2025reviewedgelargelanguage}, indicate that SLMs often suffer from degraded response quality and increased susceptibility to hallucinations—generating factually incorrect content \citep{xu2023llmcadfastscalableondevice}. This trade-off between efficiency and performance presents a critical barrier, particularly for applications requiring high accuracy, such as medical data analysis or financial processing \citep{wang2024modelcompressionefficientinference}.

To mitigate this trade-off, numerous studies have introduced approaches that dynamically route input queries to models of varying sizes, aiming to lower inference costs without compromising output quality \citep{kou2024cllmsconsistencylargelanguage, anagnostidis2024dynamiccontextpruningefficient}. Collaborative decoding has emerged as a promising approach \citep{shen2024collabdecode,shi2024inferflowefficienthighlyconfigurable}. This paradigm involves SLMs handling the bulk of the inference process while LLMs assist in generating critical tokens, such as those with high uncertainty or decisive impact on the output. Research suggests that this method leverages the strengths of both model types, maintaining efficiency while enhancing quality. For instance, Wang et al.'s study on Fast and Slow Generating (FS-GEN) \cite{zhang2024fastslow} categorizes LLMs as System 2 (slow and deliberate) and SLMs as System 1 (fast and intuitive), finding that collaborative interactions require less than 20\% of the computations, following scaling laws based on parameter ratios.
 
Building on these insights, we introduce a novel token-level routing inference system for edge devices, addressing the challenge of balancing efficiency and performance in resource-constrained settings. The system enables on-device SLMs to perform primary decoding while selectively routing critical tokens to a cloud-based LLM using a lightweight, confidence-based MLP router (See Figure \ref{fig:system_overview} for details). Empirical results on CommonsenseQA demonstrate that routing only 7\% of tokens to the LLM yields over 60\% accuracy improvement, with more than 80\% cost reduction compared to full LLM inference. This system paves the way for practical, low-latency, high-quality language model applications on edge hardware, as it mitigates the traditional trade-off between model size and performance, opening new possibilities for deploying high-quality language models in resource-constrained environments. For example, in privacy-sensitive scenarios like medical data analysis, on-device inference reduces data transmission, protecting user data, while cloud-based LLM assistance ensures accuracy.

 Unlike prior works which focus solely on routing algorithms, our contribution lies in building a fully operational client-server token routing system compatible with edge deployment. This includes integration with ONNX inference on laptops and phones, low-latency LLM serving, and practical routing logic—bringing theoretical ideas into real-world applications.

\section{Token Level Routing}
\begin{figure*}[t]
  \centering
  \includegraphics[width=0.9\textwidth]{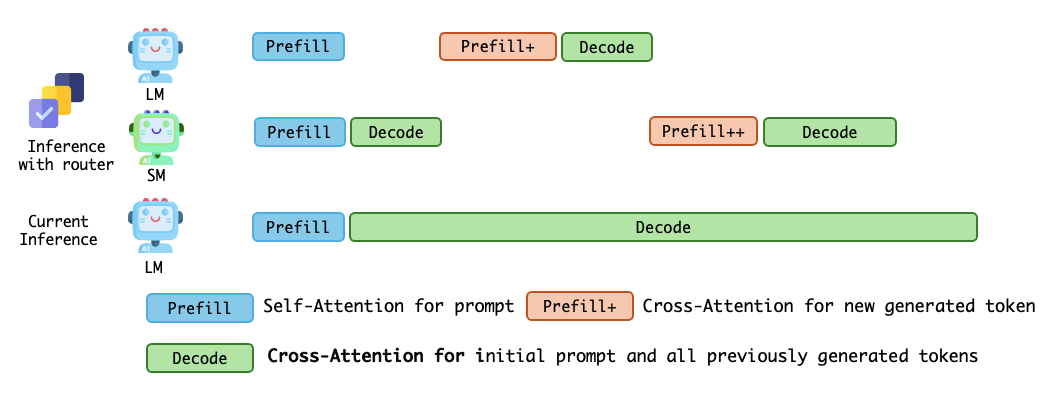}
  \caption{Computation procedure: Unlike conventional inference, the token routing system involves multiple rounds of prefill and decode within a single request, which prevents full utilization of inference acceleration engines such as SGLang and vLLM, as they only optimize kernel and KV cache on single stage prefill and decode.}
  \label{fig:computation}
\end{figure*}
In this section, we introduce serveral token level routing algorithm that can be used on our system.

\subsection{\textsc{CITER} – Collaborative Inference with Token-level Routing}

\textsc{CITER} \citep{zheng2025citer} is a framework that accelerates language model inference through token-level routing between a small, fast but less accurate language model (SLM) and a large, accurate but expensive model (LLM). A trainable router determines, for each token, whether to use the SLM or the LLM, based on routing scores and a predefined threshold $\tau$.

To capture the long-term tradeoff between cost and quality, \textsc{CITER} formulates router training as a preference-based reinforcement learning problem over a Markov Decision Process (MDP). Each state consists of the input prompt and the current generated tokens, and the actions correspond to choosing either the SLM or LLM to generate the next token. Rewards reflect both inference efficiency and the quality of the final generated response.

Rather than specifying explicit reward functions, \textsc{CITER} leverages pairwise routing preferences: whether generating a token with the SLM is preferred over the LLM. These preferences are modeled using the Bradley-Terry model and optimized via a cross-entropy loss on the routing policy. To assign token-level preferences efficiently, a shortcut mechanism is introduced. If the SLM correctly predicts the next ground-truth token, it is preferred; otherwise, if the LLM predicts it correctly, the LLM is preferred. Only when both fail is a full generation trajectory used to assess quality—drastically reducing the need for expensive full-sequence rollouts.

The router is trained iteratively. In each round, the current policy generates routing decisions to collect updated preferences, which are then used to refine the routing policy. During inference, the router deterministically selects the model based on the posterior policy $\pi(a|\mathbf{s})$, adjusted by a prior $(\rho(a_S), \rho(a_L))$, allowing flexible control of the accuracy-efficiency tradeoff via a tunable threshold $\tau = \rho(a_L)$. This enables efficient collaborative inference that maintains high response quality while substantially reducing inference cost.

\subsection{\textsc{Co-LLM} – Learning to Defer and Collaborate Efficiently}

\textsc{Co-LLM} \citep{shen2024collabdecode} is another token level routing framework that jointly updates the base model and the deferral policy by minimizing the negative log marginal likelihood of the training data. To facilitate training, an initialization scheme is introduced based on weak supervision: token-level pseudo-labels $\hat{Z}_t$ indicate whether the assistant model predicts the ground-truth token better than the base model. This initialization helps the base model quickly identify difficult tokens suitable for deferral, which are then refined via unsupervised learning.

At inference time, a threshold $\eta$ governs the deferral frequency: if $P_\theta(Z_t = 1 \mid X_{<t}) > \eta$, the base model defers to the assistant. This decoding strategy supports fine-grained, token-level control of collaboration, yielding improved performance on tasks requiring domain expertise or complex reasoning. Empirical results show that \textsc{Co-LLM} not only surpasses single-model baselines but also outperforms other multi-model strategies, while requiring significantly fewer calls to large models during inference.

\section{System Overview}

In the token routing system, we decompose the architecture into three primary modules: (1) a server-side large language model (LLM) serving module, (2) an on-device small model inference module, and (3) a token routing selection module. This system introduces a novel serving paradigm wherein a single request may involve multiple rounds of prefilling, as illustrated in Figure~\ref{fig:computation}. Crucially, interference can arise between the prefilling and decoding phases. While mainstream serving engines offer flexible separation strategies via dynamic partitioning (DP), they are not optimized for scenarios involving multiple alternating prefilling and decoding stages. Consequently, our system requires new strategies for kv-cache management and resource allocation to support efficient inference under this setting. Therefore, our goal in developing this system is to build a prototype of the token routing framework and optimize it based on its unique computational characteristics.

On the server side, we adopt SGLang \citep{zheng2024sglangefficientexecutionstructured} as our LLM serving engine due to its flexible operator definitions and extensible kv-cache management capabilities, which make it well-suited for the optimization techniques we propose. For on-device inference, existing solutions already enable the efficient deployment of small models. However, token routers—such as the routing module in \textsc{CITER} or the deferral mechanism in \textsc{Co-LLM}—often involve substantial computation. Since routing decisions must also be executed on mobile devices, we employ the ONNX \citep{onnxgithub} framework, which supports both model inference and router execution in a unified and lightweight environment. In the following demonstration and evaluation, we exclusively adopt CITER, as its MLP-based router is more amenable to deployment on edge devices.

\subsection{Front End}

\begin{figure}[h!]
  \centering
  \includegraphics[width=0.5\textwidth]{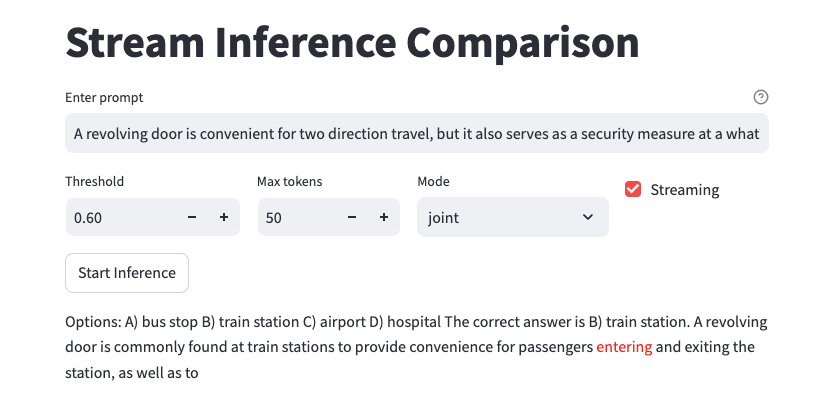}
\caption{User interface of the token-level routing system. Users can set prompts, thresholds, and decoding modes. Tokens from the large model are highlighted in red for interpretability.}
\label{fig:Front end}
\end{figure}

We design a user-facing interface to support dynamic inference under a token-level routing framework. The interface includes a \textit{prompt input field} for specifying the initial query, and a \textit{threshold slider} that governs the routing decision between the small and large models. The threshold corresponds to the confidence score predicted by an MLP classifier, which operates on the last-layer hidden state of the small model. A token is routed to the large model if its score falls below the specified threshold, reflecting insufficient confidence in the small model’s prediction.

The interface supports two inference modes: \texttt{joint}, which enables collaborative decoding between the small and large models via token-level routing; and \texttt{small\_only}, which disables routing and uses only the small model for decoding. For interpretability, tokens generated by the large model are highlighted in red during generation, allowing users to visualize routing behavior in real time.

\subsection{API CALL}

Since \textsc{CITER} requires the last-layer hidden states of the model as input to the MLP router, we design a custom API schema (See Figure \ref{fig:api_format}) to ensure that each invocation of the large language model includes the necessary internal state information. This allows token-level routing decisions to be made based on contextual representations while maintaining stateless communication across modules.

\subsection{Backend}
\begin{figure}[h!]
  \centering
  \includegraphics[width=0.5\textwidth]{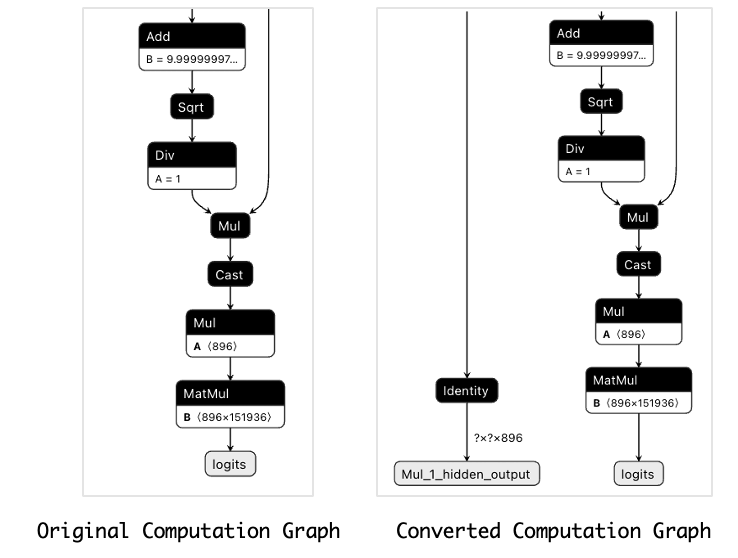}
  \caption{Left: ONNX computation graph of the original Qwen-0.5B model. Right: Modified graph with last-layer hidden states exposed as an output.}
  \label{fig:backend}
\end{figure}

On the server side, we adopt SGLang as the inference engine to serve large language models. For on-device execution, we deploy models in the ONNX format to enable lightweight and efficient inference. However, since the router requires access to the last-layer hidden states of the model to determine whether a token should be routed, we modify the ONNX model accordingly (See Figure \ref{fig:backend}). Specifically, after loading the model, the backend parses the computational graph to automatically identify the computation node corresponding to the last-layer hidden states, and programmatically registers it as an additional output.

In cases where automatic matching fails, the node name can be manually identified using tools such as Netron, and the model modification script can be invoked to transform the original ONNX model into a format compatible with the routing system.

\lstset{
  basicstyle=\ttfamily\small,
  backgroundcolor=\color{gray!5},
  frame=single,
  breaklines=true,
  columns=fullflexible
}

\begin{figure}[t]
\centering
\begin{minipage}{0.9\linewidth}
\begin{lstlisting}[language=json]
{
  "context": "The mitochondria is the powerhouse of the",
  "current_token": "cell",
  "token_index": 15,
  "routing_threshold": 0.7,
  "slm_state": {
    "hidden_states": [...],
    "attention_states": [...]
  },
  "llm_state": null,
  "history": {
    "previous_decisions": [
      {"token": "mitochondria", "route": "SLM"},
      {"token": "powerhouse", "route": "LLM"}
    ]
  },
  "meta_data": {
    "session_id": "session123",
    "request_id": "req456"
  }
}
\end{lstlisting}
\end{minipage}
\caption{An example of the custom API format used to pass internal model state and routing metadata between modules.}
\label{fig:api_format}
\end{figure}

\section{System Evaluation}

As a routing system between a small and a large model, the overall system throughput is jointly influenced by the small model's inference speed, the number of routed tokens, the communication latency between the mobile device and the server, and the backend serving system's workload. Meanwhile, the quality of the user response is ensured by the router. Therefore, we evaluate our token routing system from both a system-level perspective and a response quality perspective. We use a MacBook Pro with an M1 chip as the edge device and run the Qwen/Qwen2.5-32B-Instruct model on two A100 GPUs with the SGLang inference backend, configured with tensor parallelism (\texttt{--tp=2}). Even though onnx provide internal acceleration kernel for M1 chip, we only use CPU for small model and Router inference to simulate other edging device that do not support onnx acceleration kernel. 

\subsection{System Throughput}
\begin{table*}[ht]
\centering
\caption{Performance Metrics (in seconds) under Different Thresholds -- Non-Stream Inference}
\label{tab:performance_metrics_transposed_non_stream}
\begin{tabular}{|l|c|c|c|c|c|c|c|c|}
\hline
\textbf{Threshold}                       & \textbf{0.40} & \textbf{0.50} & \textbf{0.60} & \textbf{0.70} & \textbf{0.72} & \textbf{0.76} & \textbf{0.80} & \textbf{0.90} \\
\hline
\textbf{Routing Number}                 & 0             & 0             & 1             & 14            & 17            & 38            & 65            & 76            \\
\textbf{SLM Inference Time (s)}         & 28.19         & 28.10         & 28.40         & 28.04         & 27.58         & 27.59         & 28.02         & 28.20         \\
\textbf{TTFT (s)}                        & 0.67          & 0.50          & 0.45          & 0.34          & 0.46          & 0.41          & 0.47          & 0.49          \\
\textbf{TBT for SLM (s)}                & 0.28          & 0.28          & 0.28          & 0.33          & 0.33          & 0.45          & 0.80          & 1.18          \\
\textbf{Comm + LLM Inference (s)}       & 0.00          & 0.00          & 0.94          & 11.97         & 13.43         & 34.00         & 58.23         & 72.76         \\
\textbf{Overall (s)}                    & 28.14         & 28.15         & 28.40         & 40.06         & 41.30         & 61.65         & 86.32         & 101.05        \\
\hline
\end{tabular}
\end{table*}

\begin{figure*}[ht]
\centering
\subcaptionbox{Communication + LLM Inference Time\label{fig:comm_llm}}[0.3\linewidth]{%
    \includegraphics[width=\linewidth]{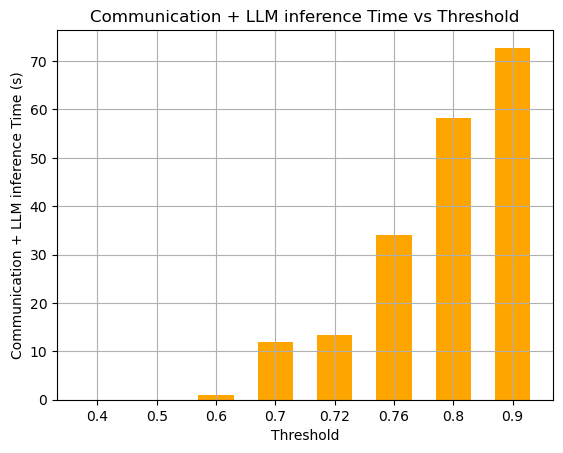}
}
\hfill
\subcaptionbox{Complete Request Time\label{fig:complete_request}}[0.3\linewidth]{%
    \includegraphics[width=\linewidth]{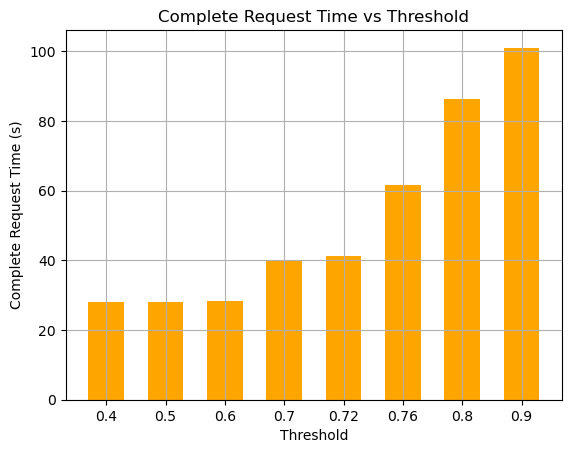}
}
\hfill
\subcaptionbox{Time Between Tokens for SLM\label{fig:tbt}}[0.3\linewidth]{%
    \includegraphics[width=\linewidth]{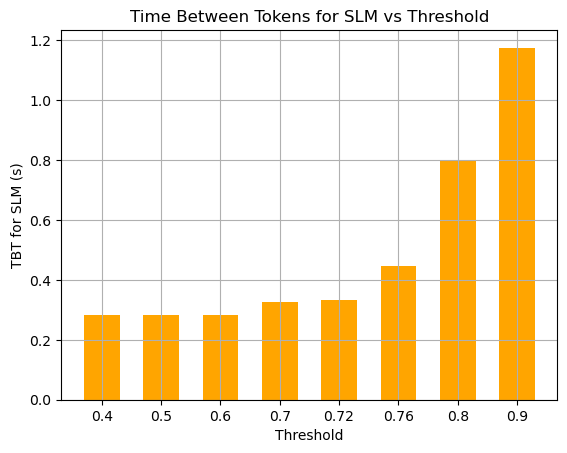}
}
\caption{Latency comparisons under different thresholds.}
\label{fig:latency_comparison}
\end{figure*}

In our evaluation, we randomly selected 100 multiple-choice questions from the CommonsenseQA dataset. For each inference, the maximum generation length was set to 100 tokens. We varied the threshold of the MLP-based router from 0.4 to 0.9, where the threshold determines the routing score required for a token to be forwarded to the large language model (LLM).

Table \ref{tab:performance_metrics_transposed_non_stream} shows the streaming and non-streaming inference speed of our system. The time to first token (TTFT) reflects the prefill time of the SLM. When the threshold is low, all tokens are generated locally by the SLM, which achieves an average generation speed of approximately 4 tokens per second on an M1 chip. When the threshold reaches 0.3, the router begins forwarding some tokens to the LLM for inference.

To simulate a worst-case deployment scenario, we assume a network communication delay of approximately 170 milliseconds between the client and server. Each LLM request incurs a latency of around 0.9 seconds. Furthermore, transferring the generation context from the LLM back to the SLM introduces an additional prefill delay of approximately 4 milliseconds, which accumulates as the number of LLM calls increases.

As the number of routing events increases, the time between tokens (TBT) begins to rise accordingly. This is primarily due to the lack of a key-value cache (kv-cache) management mechanism in the current ONNX-based inference system, which necessitates re-prefilling the entire sequence during each routing operation. Consequently, this leads to increased latency. Under more favorable network conditions—such as scenarios where edge devices maintain direct connections to the server—the system is expected to exhibit significantly improved performance.

\subsection{Response Eval}

Since the number of times the large language model (LLM) is involved in the inference process directly affects the quality of the final response, this section evaluates the performance of the token routing system on the CommonsenseQA dataset under various threshold settings. It is worth noting that the LLM and SLM used in the CITER \citep{zheng2025citer} were Qwen2-72B and Qwen2-1.5B, respectively. However, due to the relatively slow inference speed of the 1.5B model on edge devices, we adopt a different configuration in our routing system to ensure a better user experience. Specifically, we use the Qwen2.5-32B model as the serving LLM and the Qwen2.5-0.5B model for on-device inference, thereby achieving higher overall system throughput.

\begin{figure}[h!]
  \centering
  \includegraphics[width=0.45\textwidth]{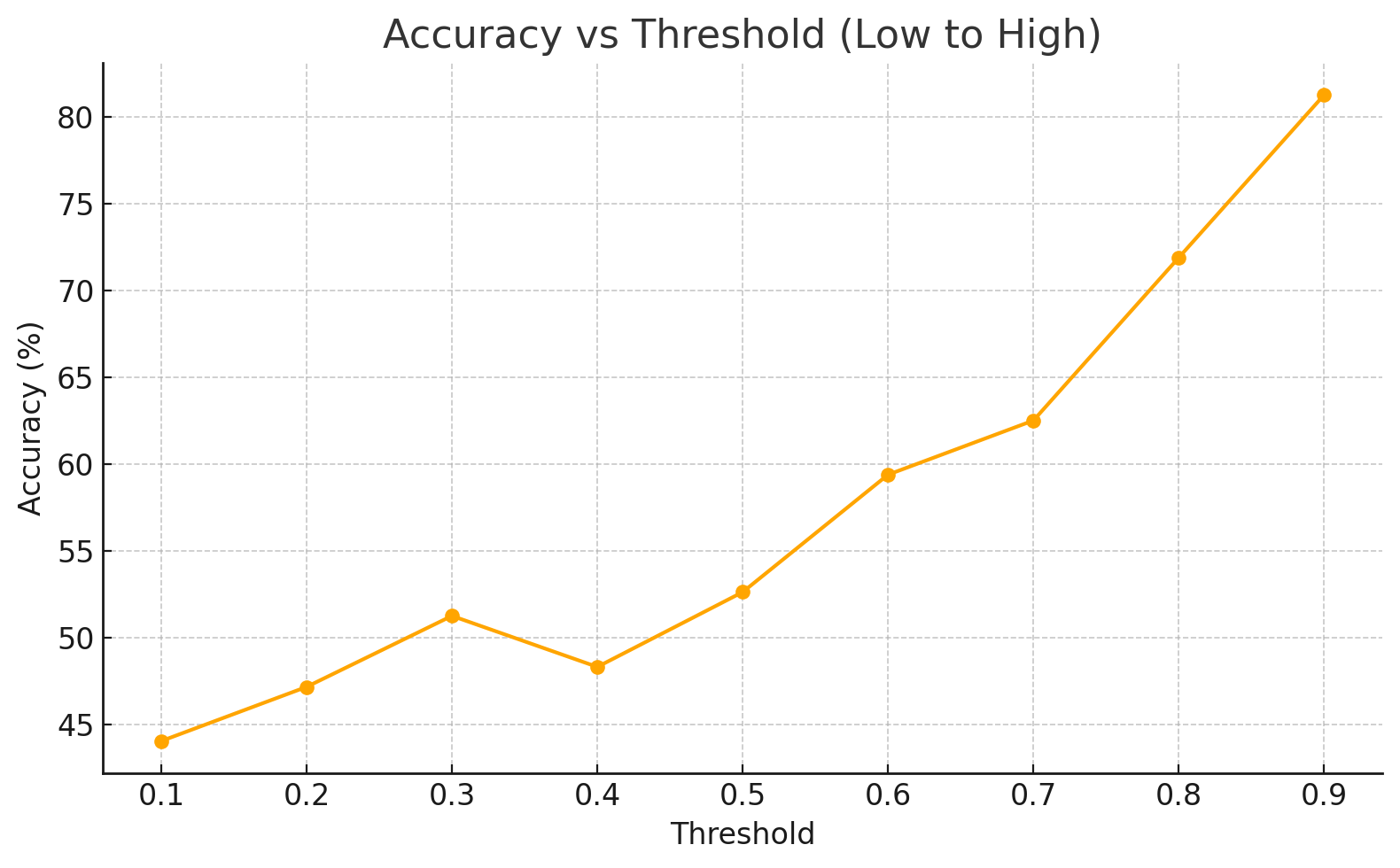}
  \caption{Accuracy vs Threshold on CommonSense QA}
  \label{fig:accuracy}
\end{figure}

\begin{figure}[h!]
  \centering
  \includegraphics[width=0.45\textwidth]{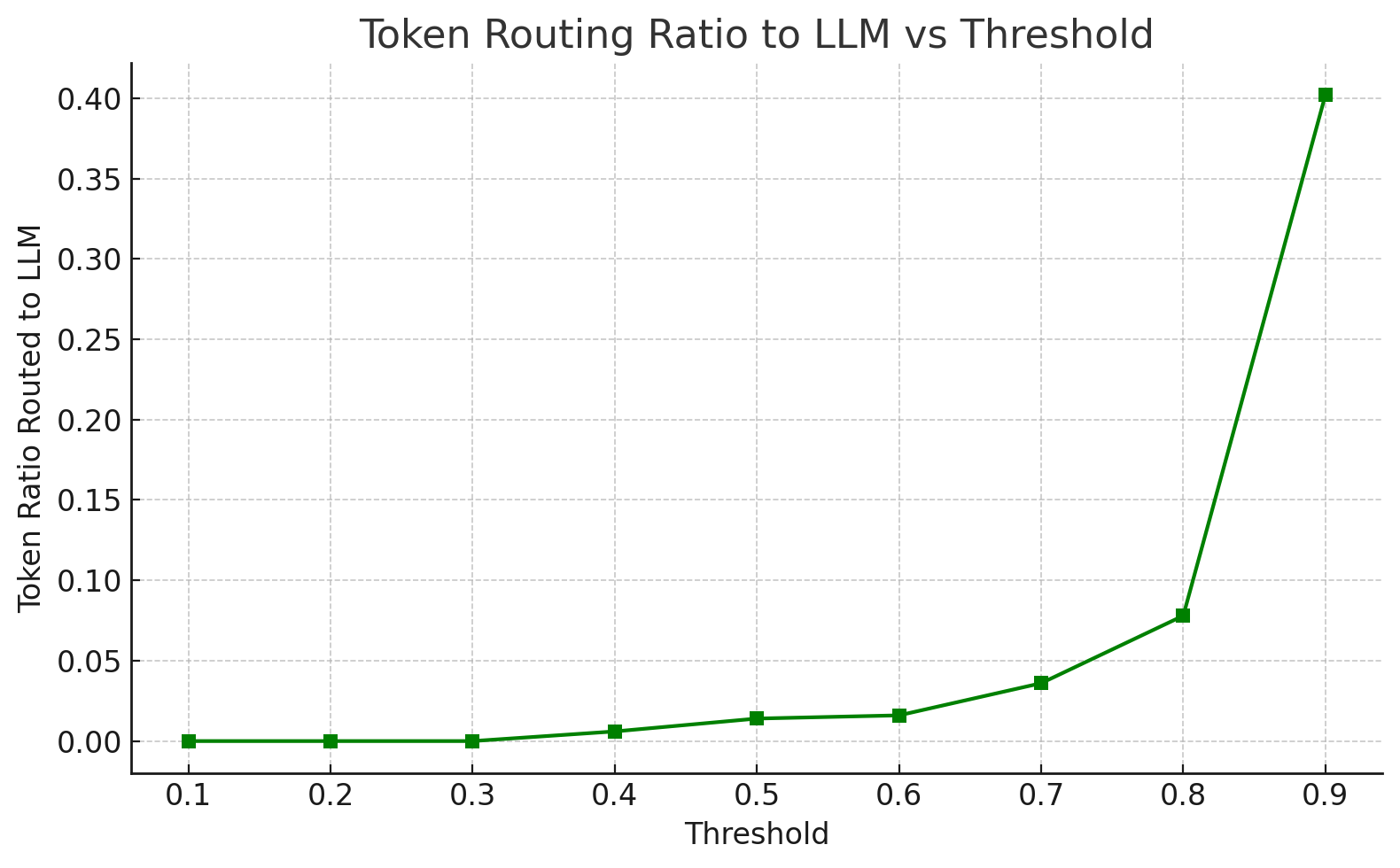}
  \caption{The ratio of tokens routed to LLM vs Threshold on CommonSense QA}
  \label{fig:Ratio}
\end{figure}

We evaluated the system performance on the CommonsenseQA dataset under various threshold settings. As shown in Figure \ref{fig:accuracy} and Figure \ref{fig:Ratio}, when the threshold falls below 0.3, the responses are predominantly generated by the small model, resulting in an accuracy of approximately 50\%, which is significantly higher than the random guess baseline of 20\%. As the threshold increases beyond 0.4, a portion of the tokens begins to be routed to the large model for decoding, leading to improved answer quality. To strike a balance between response quality and system efficiency—avoiding excessive latency introduced by frequent large model invocations—we typically set the threshold between 0.7 and 0.8 for commonsense reasoning tasks.

\section{Conclusion}

Building upon the token routing algorithm, we design a cloud-assisted token routing system that operates on devices running lightweight models at the edge. By routing a small subset of critical tokens to a large-scale model in the cloud for inference, the system significantly enhances the performance of the edge model while maintaining low inference latency. This architecture is well suited for scenarios where on-device deployment is required but model performance cannot be heavily compromised. Our experiments demonstrate that, on the CommonsenseQA dataset, routing merely 7\% of the tokens to the large model yields over a 60\% improvement in the small model’s accuracy.

\newpage

\bibliography{custom}

\end{document}